\documentclass[11pt,a4paper]{article}
\usepackage[hyperref]{acl2018}
\usepackage{times}
\usepackage{latexsym}
\usepackage{latexsym}
\usepackage{amsmath,amssymb}
\usepackage{tabularx}
\usepackage{multirow}
\usepackage{arydshln}
\usepackage{graphicx}
\usepackage{caption}
\usepackage{subcaption}
\usepackage{color}
\usepackage{listings}
\usepackage{url}
\newcommand{\specialcell}[2][c]{\begin{tabular}[#1]{@{}c@{}}#2\end{tabular}}

\aclfinalcopy % Uncomment this line for the final submission
\title{Sequence to Logic with Copy and Cache}

\author{Javid Dadashkarimi \\
   Computer Science Department\\
  Yale University\\
  {\tt javid.dadashkarimi@yale.edu} \\\And
  Sekhar Tatikonda \\
  Department of Statistics and Data Science  \\
  Yale University \\
  {\tt sekhar.tatikonda@yale.edu } \\}

\date{}

\begin{document}
\maketitle
\begin{abstract}
  Generating logical form equivalents of human language is a fresh way to employ neural architectures where long short-term memory effectively captures dependencies in both encoder and decoder units. 
  The logical form of the sequence usually preserves information from the natural language side in the form of similar tokens, and recently a copying mechanism has been proposed which increases the probability of outputting tokens from the source input through decoding.
  In this paper we propose a caching mechanism as a more general form of the copying mechanism which also weighs all the words from the source vocabulary according to their relation to the current decoding context.
  Our results confirm that the proposed method achieves improvements in sequence/token-level accuracy on sequence to logical form tasks.
Further experiments on cross-domain adversarial attacks show substantial improvements when using the most influential examples of other domains for training.
\end{abstract}

\section{Introduction}
Understanding human language has a long history, and interpreting language for machine execution is a key component of artificial intelligence and machine learning \cite{Dong:2016,Jia:2016}.
Multiple techniques have been proposed in a  neural setting for text to text translation \cite{Bahdanau:14,Sutskever:14,Luong:15} or in machine format \cite{Jia:2016,Zhong:2017,Quirk:15} including, but not limited to, semantic parsing \cite{Jia:2016}, code generation \cite{Quirk:15}, and query generation for database systems \cite{Xiaojun:17,Zhong:2017}.

Recently \citet{Jia:2016} proposed an augmented pointer network for converting a sequence to its logical form where a greedy decoder generates the next token according to the last hidden state of the source encoder augmented by a pointer to each of the source tokens. 
This is a very straightforward mechanism to pay attention to the source sequence which increases the chance of generating shared tokens between the natural language input and its logical form output.
In this paper we introduce a ``cache" to take the entire source vocabulary into account during decoding. 
The following example is from the GeoQuery data set \cite{Tang:2001,Zettlemoyer:2012}:

\begin{itemize}
    \item[-] \texttt{x: what rivers flow through colorado ?}
    \item[-] \texttt{y: answer ( A , ( river ( A ) , traverse ( A , B ) , const ( B , stateid ( colorado ) ) ) )}
\end{itemize}

The previously proposed augmented pointer network increases the probability of \texttt{colorado} in the output logical form sequence as this entity appears in the input sentence.
However, our proposed joint cache distribution aims to find semantically related tokens throughout the entire source vocabulary which hold importance in outputting relevant tokens. 
%  TODO this is good, although I still don't like the sentence "The reason is that sequence level ..." It could be clarified -> done

For this example, the caching probability for the words `\texttt{flow through}' turns on in this context and generates `\texttt{traverse}' from the probability distribution over the source vocabulary.
We note that the previous model proposed by \citet{Jia:2016} is expected to do the same through a general decoder trained over the data set. 
Our findings reveal that in some situations neither the local copy nor the general decoder are able to find in-domain vocabulary words (e.g., \texttt{date} for the calendar domain or \texttt{author} for the publications domain).
Our proposed method on the other hand is able to capture these situations in different domains. While this method provides varied tokens and improves the token-level accuracy, similar to the baselines, it is not able to boost the sequence level accuracy significantly and provides marginal improvements in this metric.

In the second part we use cross-domain adversarial attack to augment our model with cached tokens in other domains.
To do this we follow the influence functions used in \cite{Koh:2017} to find the most sensitive examples.
Our results support the usefulness of these examples compared to the original version of training data.

% Although the current data sets are not specific to domains which inherently repeat words from the source vocabulary, the proposed model achieves consistently better performance for token-level accuracy compared to the state-of-the-art sequence to logic models using fixed settings.

\section{Related Work}
\subsection{Sequence to logic}
Semantic parsing maps natural language sentences to formal meaning representations. Semantic parsing techniques can be performed on various natural languages as well as task-specific representations of meaning. A semantic parser can be learned in a supervised or semi-supervised manner in which the natural language sentence is paired with either a logical form or its executed denotation. 

\textbf{Supervised Semantic Parsing}
In recent work on supervised semantic parsing, a natural language statement is paired with its corresponding logical meaning representation \cite{Zettlemoyer:05,Wong:06,Kwiatkowksi:10}. 
The logical forms may be database queries, dependency graphs, lambda-calculus terms, among others. 
% A common approach has been to induce semantic parsers from data with the use of a probabilistic grammar paired \cite{Zettlemoyer:05}.
% Commonly used is the combinatory categorical grammar (CCG) \cite{Steedman:00}, which consists of a language-specific lexicon that pairs words and phrases with both syntactic and semantic information and a set of combinatory rules. CCG's interweaving of syntax and semantics makes it powerful for semantic parsing \cite{Zettlemoyer:05,Zettlemoyer:07,Kwiatkowksi:10,Kwiatkowski:11, Zettlemoyer:05, Zettlemoyer:07}. 

% Additionally, recent work in deep learning has led to neural semantic parsing which aims to directly translate natural language to logical forms without using hand-engineered features or inducing a grammar.  \cite{Dong:2016,Jia:2016}.  \cite{Dong:2016} use a logical-form agnostic sequence to sequence model but also propose a tree-based decoder to deal with the nested nature of logical forms. \cite{Jia:2016} introduce an augmented pointer network to copy the source tokens to the final output layer.

\textbf{Semi-Supervised Semantic Parsing}
Due to the lack of large supervised datasets, semi-supervised semantic parsing is often performed on denotations from question-answer pairs, which are much cheaper to obtain than corresponding logical forms.
\citet{Liang:13} map questions to answers using latent logical forms. \cite{Cheng:17} use a transition-based approach to convert natural language sentences to intermediate predicate-argument representation structures then grounded to a knowledge base. \citet{Kwiatkowski:13} build a logical-form meaning representation which is fed to an ontology matching model. 
% Additionally,\cite{Reddy:14} view the task as a graph-matching problem between the grounded CCG parsed semantic graph and a knowledge base. 
For this paper, however, we focus on neural supervised approaches to semantic parsing. 

\section{Sequence to Logic with Copy and Cache}
\label{Sequence to Logic with Copy and Cache}
Let $y^t=\textrm{argmax}(\dot{\beta})$ be the next token to be generated at time stamp $t$ in a standard sequence to sequence decoding process. We refer the reader to \cite{Jia:2016} for the equations upon which we base ours below. As in \cite{Jia:2016} we create a distribution over the target vocabulary as well as a copying distribution. 
However, our copying mechanism constitutes both a distribution over the source sentence and the entire source vocabulary combined with the target distribution through the concatenation in Equation \ref{eq:e_ji}:

\begin{flalign}
&e_{ji} = [s_j^TW^{(a)}b_i] + [z_t\odot f(s_j,c_j)]\label{eq:e_ji} \\
% f(s_j)=s_j^TW^{(h)} \label{eq:e_ji}
&\alpha_{ji} = \frac{\exp(e_{ji})}{\sum_{i'=1}^m \exp (e_{ji'})} \\
&c_j = \sum_{i=1}^{|V_s|} \alpha_{ji}b_i \\
&P(y_j=w|x,y_{1:j-1}) \propto \exp (U_w[s_j,c_j]) \\
&P(y_j=\textrm{copy}[i]|x,y_{1:j-1}) \propto \exp(e_{ji}) 
\label{eq:copy_dist}
\end{flalign}
where $e_{ji}$ is the normalized attention score of state $s_j$ to the source annotation $b_i$; $w$ is the target candidate token and copy[$i$] is the candidate token to be copied from the source sequence.
We refer to $f(s_j,c_j)$ in this equation as the cache function, as it aims to keep track of the history with which words in the input vocabulary and input sequence appear in contexts relevant to the target output. 
One simple approximation for $f(s_j,c_j)$ is $f(s_j,c_j) = s_j^TW^{(a)}W^{(h)}$ where $W^{(a)} \in \mathbb{R}^{d\times 4d}$ and $W^{(h)} \in \mathbb{R}^{4d\times |V_s|}$ is the cache matrix (see the supplementary material for more examples of this function). 
It's worth noting that since we select the bi-directional LSTM for our encoding annotations, $W^{(a)} \in \mathbb{R}^{d\times (\overrightarrow{2d}+\overleftarrow{2d})}$, where $d$ is the hidden layer size and $2d$ results from the combination of the $h$ and $c$ parameters of the LSTM model. 
% We experiment with multiple history functions which vary in their nonlinearities, history matrix combinations as well as reset gates.
% TODO explain why the reset gate -- to answer that one reviewer -> done
$z_t=\sigma{([s^T_jU_{z_t}]+[c^T_jW_{z_t}])}$ is a reset gate for the context $c_j$ and state $s_j$ where $U_{z_t},W_{z_t} \in \mathbb{R}^{d\times |V_s|}$. This is inspired by the forget and reset gates in LSTM's and Gated Recurrent Units.
As the history matrix calculates often a large distribution, this distribution may be noisy. 
This reset gate allows us to diminish the effect of irrelevant words, and 
we expect that $W^{(h)}$ will be learned over the source vocabulary. 
\begin{figure}[h]
  \centering
    \includegraphics[width=0.85\linewidth]{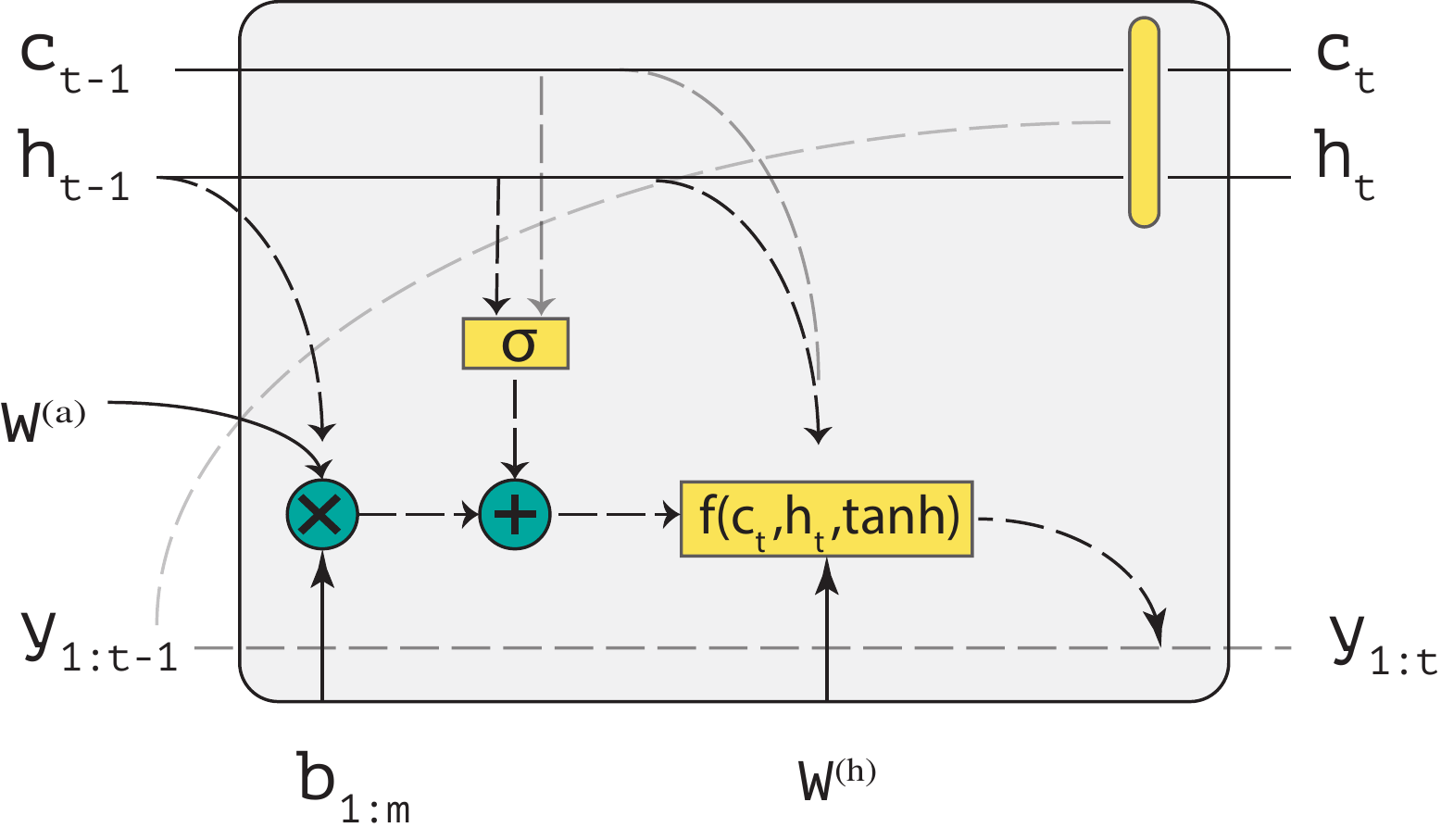}
     \caption{A caching gate for decoding. For prediction we use both the local attention $W^{(a)}$ and the caching matrix $W^{(h)}$. }
\label{fig:hgate}
\end{figure}
Fig. \ref{fig:hgate} shows the sketch of the proposed caching mechanism. The remaining equations amount to a standard attention mechanism with our additional caching mechanism analogous to the copying equations in \cite{Jia:2016}.

\textbf{Influence functions}
Recently \citet{Koh:2017} introduced influence functions for a number of classification tasks to find the most influential examples for testing.
The central goal of these functions is the non-parametric estimation of $\hat{\theta}_{\epsilon,\textbf{z}} = \arg\min_{\theta}\sum_{z_i \notin \textbf{z}}\frac{1}{n}\mathcal{L}(z_i,\theta) + \epsilon \mathcal{L}(\textbf{z},\theta,t_k)$  (i.e., parameters to learn) by up-weighting $\forall z_i\in \textbf{z}$ (i.e., batch of examples $z_j=(x_j,y_j)$) or
$\hat{\theta}_{\epsilon,z_j} = \arg\min_{\theta}\sum_{z_i\neq z_j}\frac{1}{n}\mathcal{L}(z_i,\theta) + \epsilon \mathcal{L}(z_j,\theta,t_k)$ by up-weighting one particular example $z_j$ \cite{Wasserman:2006,Cook:1980}.
Later in Section \ref{Results and Discussion} we study the impacts of other domains in caching useful words both on domain-level (i.e., $\hat{\theta}_{\epsilon,\textbf{z}}$) and example level (i.e., $\hat{\theta}_{\epsilon,z_j}$) by adversarial attacks on binary classification loss $\mathcal{L}(.)$.
Our experiments in Section \ref{Cross-domain adversarial attack} show that a close domain $\textbf{z}$ can provide useful information for predicting $y_i$.

\section{Experiments}
\subsection{Experimental Settings}
\textbf{Data sets:}
We evaluate the models based on standard semantic parsing datasets.

\textbf{GeoQuery}:
The GEO dataset \cite{Wong:06} contains natural language queries about U.S. geography and their associated Prolog queries. We use the standard split of 680 training examples and 200 test examples \cite{Zettlemoyer:05}. We follow the preprocessing of \cite{Jia:2016} and use De Bruijn index notation for variable-name standardization. 

% \textbf{ATIS}:
% The ATIS dataset contains contains natural language queries to a flights database paired with lambda-calculus queries. We use the split of 4473 training examples and 448 test used by \cite{Zettlemoyer:07}

\textbf{Overnight}:
The Overnight dataset contains natural language paraphrases paired with logical forms from eight domains such as restaurants, publications and basketball. The dataset was developed by \cite{Wang:2015} using a crowdsourcing experiment in which Amazon Mechanical Turkers created paraphrases for given logical forms which were generated from a grammar.

\textbf{Parameter settings:} We set the hidden size $|h_t|=200$ and the input embedding dimension $|x_t|=100$.
We also use Long Short-Term Memory (LSTM) units as the basic encoder/decoder unit in all of our experiments.
We initialize all the variables with a uniform distribution in $[-1,1]$ and use stochastic gradient descent as our learning method with a $0.5$ learning rate decreasing by half at each epoch.
We set the number of epochs to $30$ for all the experiments.
% We do not claim that our negative likelihood function is convex but we do expect that all the minimums are good.

In the decoding process we assume a maximum length of $100$ for all the data sets and stop decoding after predicting the end of the sentence indicator.  

\subsection{Results and Discussion}
\label{Results and Discussion}
In this part we assess the validity of the proposed network on two different tasks. 
To be fair, we fix the settings for all the runs and then test the performance over our sampled datasets.
For example, we do not use any rule-based augmented samples as in \cite{Jia:2016}, which is a domain-specific technique and boosts the performance for all the baselines.
% For sequence to SQL, \cite{Xiaojun:17} recently propose a robust network which outperforms the reinforcement-learning based algorithm by \cite{Zhong:2017}.
% This is also considered a domain specific model that potentially improves the accuracy. 
% However, as we discuss later, the general logical form models achieve higher token-level accuracy on small datasets.
% In the following sections we will discuss the results on each task separately.

\begin{table*}[tp]
\centering
\begin{tabular}{lllllll}
\cline{1-7}
    \multirow{2}{*}{ID}&\multicolumn{2}{c}{GEOQUERY}  & \multicolumn{2}{c}{GEOQUERY-S} & \multicolumn{2}{c}{OVERNIGHT}  \\ \cline{2-7} 
     &SEQ & TOK & SEQ & TOK & SEQ & TOK \\ \cline{1-7}
     copy&0.771&0.883& 0.58& 0.865 & 0.601&  0.868\\ \cline{1-7}
     copy\&cache& \textbf{0.775} & \textbf{0.901} 
     &\textbf{0.70} & \textbf{0.886} &\textbf{0.610} &\textbf{0.871}\\ \cline{1-7}
\end{tabular}
\caption{Accuracy of different semantic parsing approaches on three collections. GEOQUERY-S is the stripped version of GEOQUERY\footnote{we remove all $\_$ and replace the logic tokens by their human language representation (e.g., \_loc with location)}.}
\label{tab:res_table:seq2logic}
\end{table*}

\begin{figure*}[t]
    \centering
    \begin{subfigure}[t]{0.32\textwidth}
        \centering
        \includegraphics[width=\linewidth]{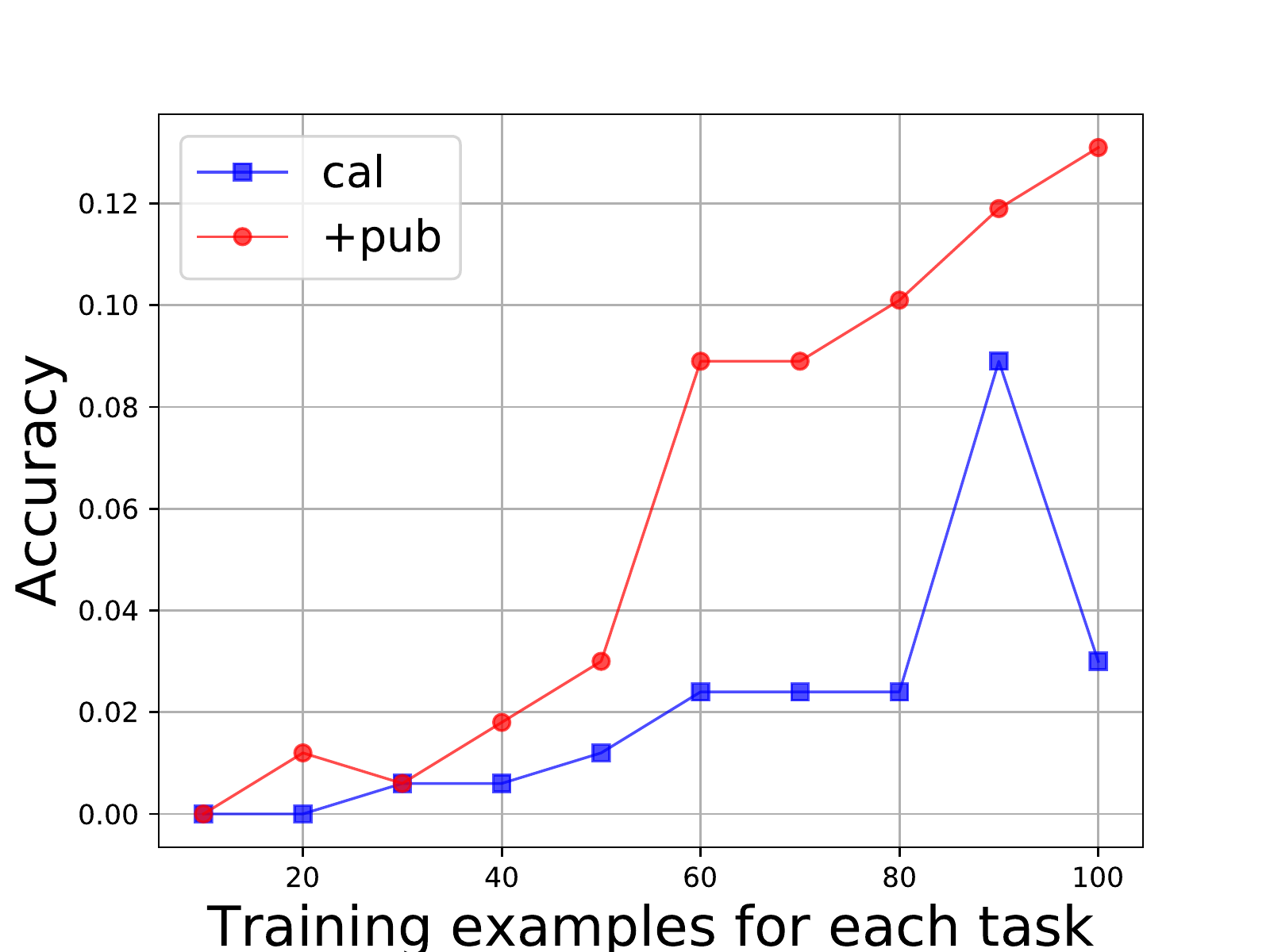}
        \caption{\texttt{pub$^*$+cal:seq}}
    \end{subfigure}
    ~
    \begin{subfigure}[t]{0.32\textwidth}
        \centering
        \includegraphics[width=\linewidth]{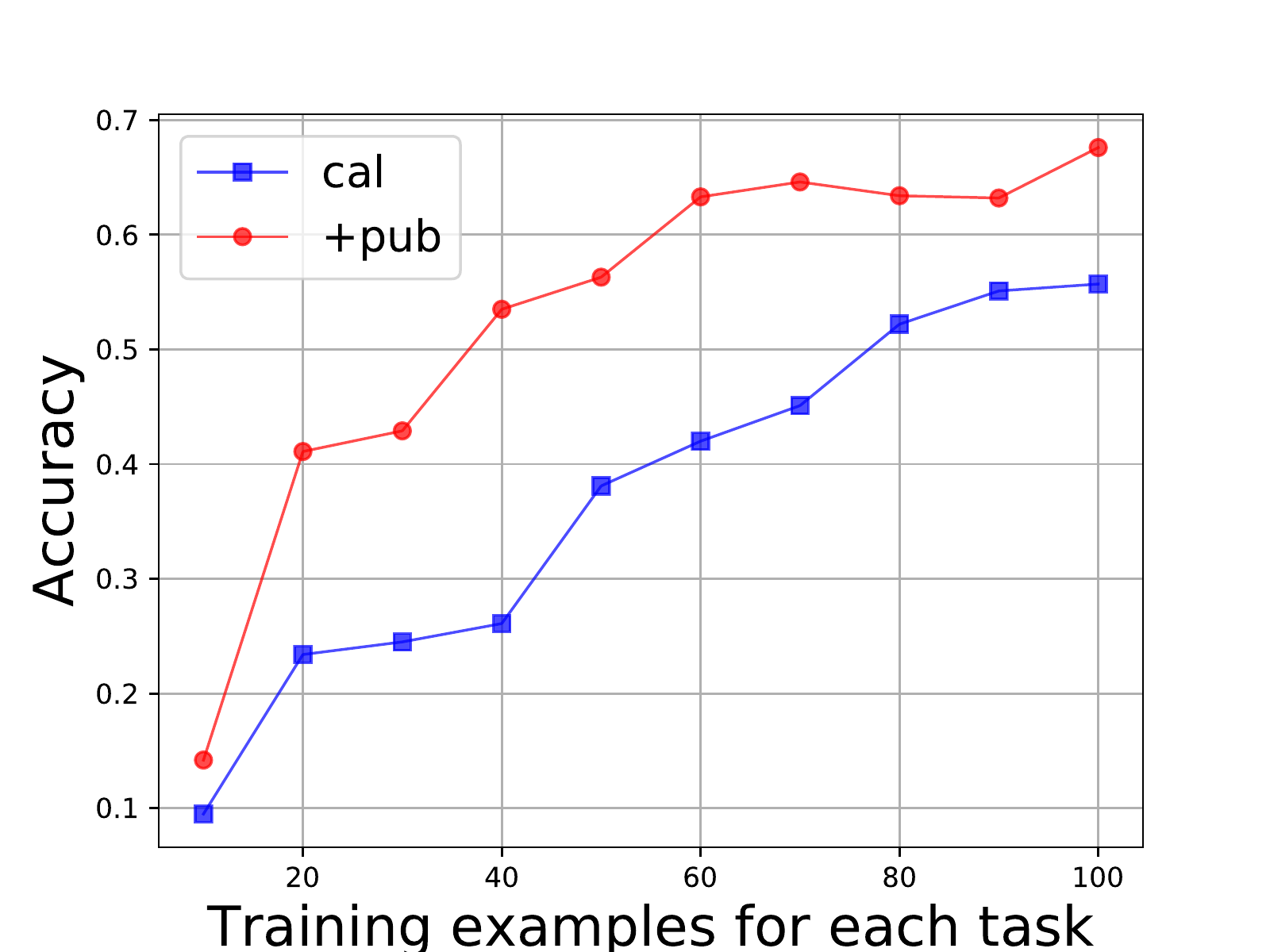}
        \caption{\texttt{pub$^*$+cal:tok}}
    \end{subfigure}
            ~
    \begin{subfigure}[t]{0.32\textwidth}
        \centering
        \includegraphics[width=\linewidth]{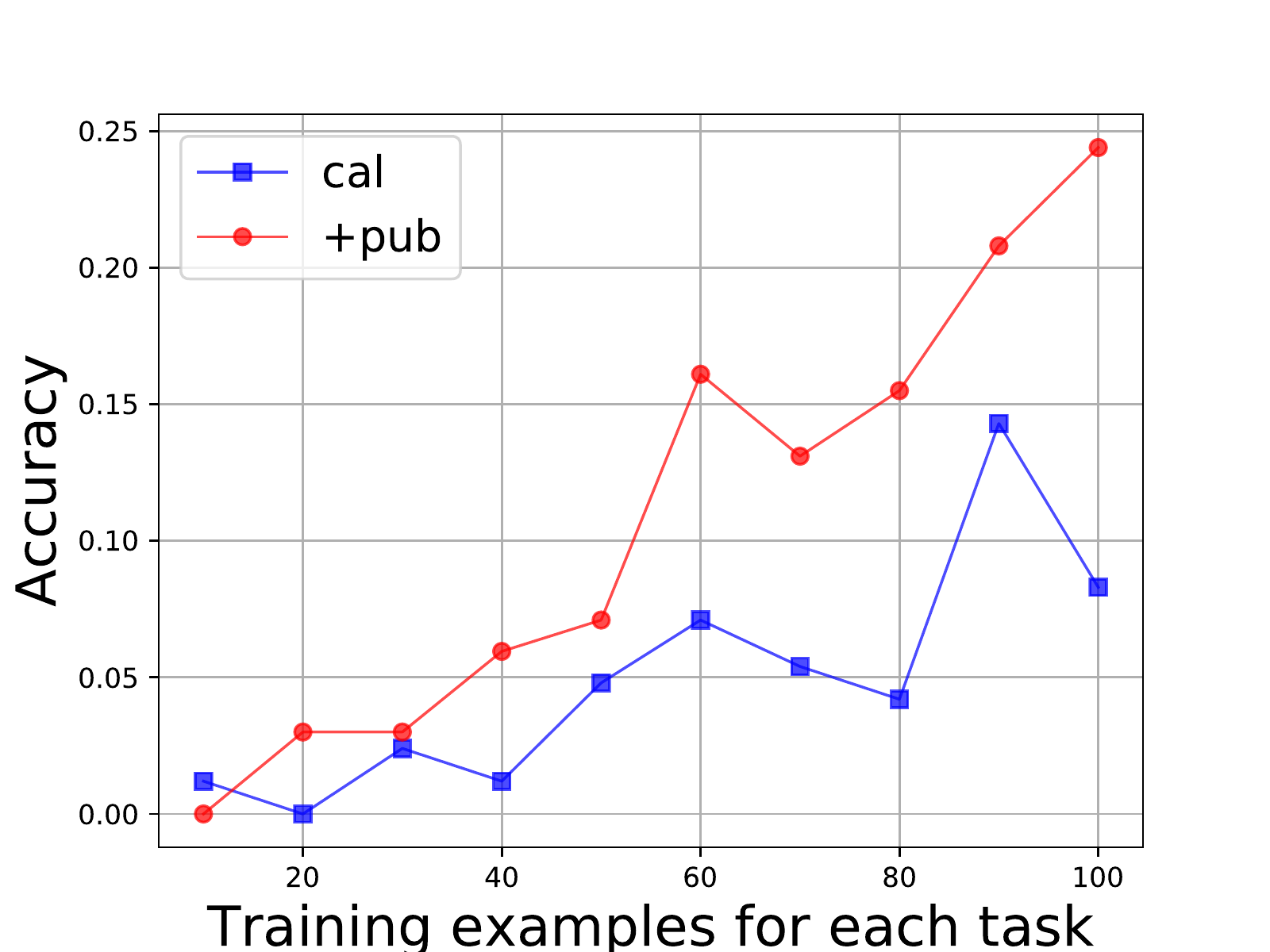}
        \caption{\texttt{pub$^*$+cal:den}}
    \end{subfigure}
    \caption{Cross-domain adversarial attack between \texttt{publications} domain and \texttt{calendar}. Results are in sequence-level (\texttt{seq}) and token-level (\texttt{tok}), and denotation-level (\texttt{den}). Auperscript $^*$ indicates that we used the most influential examples of \texttt{pub}. We add these examples at different size steps.}
    \label{fig:random-hvp}
\end{figure*}

Table \ref{tab:res_table:seq2logic} shows the experimental results on semantic parsing datasets. 
GEOQUERY-S, is a variation of the GEOQUERY dataset where logical form tokens are replaced by human language words.
% TODO clarify this sentence below 
% Therefore we limit our experiments to only the sequence and token level to match the predicted sequences.
As shown in the table, the copy\&cache algorithm works as well as the original version proposed by \cite{Jia:2016} but has further improvements in different collections.
The highest improvement is for GEOQUERY-S where it is more likely to copy a human language token to the logic form while the source sequence may have different variants of this token.
In the supplementary material,  we provided a number of examples where the original coping mechanism was not able to predict the correct logic sequence (e.g., providing semantically related words or correcting the miss-spelling tokens in the source sequence).
We also conducted a set of experiments on WikiSQL, the sequence to SQL dataset \cite{Zhong:2017} and ATIS  \cite{Zettlemoyer:07}, a flight data set, but we do not report the results since these datasets are designed for execution level comparisons in real databases and achieve comparatively lower results in sequence-level accuracy.
\footnote{In WikiSQL different SQL queries might have same results and in ATIS  a large number of operators are used instead of real tokens. \citet{Jia:2016} have not reported sequence level accuracy for ATIS either.}

\subsection{Cross-domain adversarial attack}
\label{Cross-domain adversarial attack}

In this section we experiment with a type of transfer learning to copy the cached tokens from other domains while preserving the accuracy of the running system \cite{Su:17,Fan:17}. 
First we fixed \texttt{calendar} as a test domain from OVERNIGHT and conducted several leave-one-out experiments to find the closest domains for \texttt{calendar}.
Our experiments have shown \texttt{atis} as the farthest domain from \texttt{calendar} in OVERNIGHT  and \texttt{publications} as the closest one.
We then repeated this experiment for each example of \texttt{publications} to find the most influential examples.
To this aim we used cross-domain adversarial attacks based on the inflluence functions discussed in Section \ref{Sequence to Logic with Copy and Cache}.
Then the most influential examples in the binary classification task (i.e., \texttt{calendar} and \texttt{publications}) are selected to add to our sequence to logic problem. 
To do so, we repeated the hessian vector product (HVP) algorithm used in \cite{Koh:2017} $1000$ times and created a distribution based on the number of occurrences of each example\footnote{It's worth noting that HVP is an iterative algorithm to make the example-level leave-one-out faster.}.  
Then we sampled $100$ examples for our training data.
Fig. \ref{fig:random-hvp} shows the experimental results which support the usefulness of caching from related domains. 
As shown in the figure the original \texttt{calendar} baseline over-fits to the training data (see drop in the blue line after $n=90$) while the copy\&cache version generalizes better.

\section{Conclusion and Future Work}
In this paper we introduce a caching mechanism for encoder/decoder systems.
The previous copying mechanism employed boosts performance by giving a chance for the source tokens to appear in target decoding.
The proposed function extends this possibility for the entire source vocabulary that might have some semantic relationship with the current context. 
Experiments on sequence to logic  data sets support its superiority in different evaluation metric. 
Our experiments on cross-domain adversarial attack show substantial improvements when we augment a test data by the closest examples from other domains.
We propose this method for machine translation as a possible future work where we might copy a translation of other source tokens rather than the local one. 
%\bibliographystyle{acl}
%\bibliography{acl2018}
\clearpage 
\bibliography{ref}
\bibliographystyle{acl_natbib}

\appendix
\section{Examples of cached tokens}
\label{appendix}

\begin{table*}[t]
\small
\centering
\caption{Examples from different domains where the decoder generated the target token from the caching distribution. As mentioned before there are three possibility to generate token: 1-general decoder 2- local copying, and 3- caching functions. These examples are in the third category.}
\scalebox{0.95}{
\begin{tabular}{cc} \hline
$x$ & $y$\\ \hline
\specialcell{\texttt{housing units}\\ \texttt{ with 2}\\\texttt{neighborhoods}}     
& \specialcell{\texttt{( call SW.listVal. ( call SW.countComp.}\\\texttt{ ( call SW.getProp. ( call SW.singleton } \\ \texttt{en.housing\_unit )  ( string ! type ))}\\ \texttt{( string neighborhood ) ( string = ) }\\ \texttt{( \textbf{number} 2 )))}}\\ \hline
\specialcell{\texttt{article cited by }\\\texttt{ article which }\\\texttt{in 2004}\\\texttt{is published }}     
& \specialcell{\texttt{( call SW.listValue ( call SW.filter ( call}\\\texttt{ SW.getProperty ( call SW.singleton } \\ \texttt{en.article ) ( string ! type ) )  }\\ \texttt{( call SW.reverse ( string cites ) ) ( }\\ \texttt{string = ) ( call SW.filter ( call }\\ \texttt{SW.getProperty 
( call SW.singleton}\\\texttt{ en.article ) ( string ! type ) )} \\ \texttt{ ( string publication\char`_date ) } \\\texttt{( string = ) ( \textbf{date} 2004 -1 -1 ) ) ) )}  }\\ \hline
\specialcell{\texttt{Who won the game}\\ \texttt{that took place}\\ \texttt{at stadium} \\\texttt{australia,} \\\texttt{sydney (10)?}}  &
\specialcell{\texttt{SELECT col2 FROM} \\ \texttt{table\_1\_11236195\_5 WHERE } \\
\texttt{col5='Stadium Australia ,}\\  \texttt{\textbf{Sydney} (10)'}}\\ \hline
\specialcell{\texttt{select a block} \\ \texttt{that has a width} \\ \texttt{that is the same} \\\texttt{ width of block 1 } }&
   \specialcell{\texttt{ ( call SW.listValue}\\
  \texttt{ ( call SW.filter ( call SW.getProperty}\\
  \texttt{ ( call SW.singleton en.block ) ( string ! \textbf{type} ) )} \\
  \texttt{ ( call SW.ensureNumericProperty ( string width ) ) }\\
  \texttt{ ( string <= ) ( call SW.ensureNumericEntity ( call}\\ \texttt{SW.getProperty en.block.block1} \\
  \texttt{ ( string width ) ) ) ) )}}
 \\
\hline
\end{tabular}}
\label{tab:examples}
\end{table*}

As shown in Table \ref{tab:examples}, \texttt{date} is a frequent word in the publications domain and many human languages contain similar tokens; in this particular example it has been generated from the cache function while there was a possibility to be generated from the target distribution or the copying distribution.
Indeed here this model captures general words in this domain and adds to the logical form output when there is either no signal from either human input text (i.e., $P(y_j=\textrm{copy}[i]|x,y_{1:j-1})$) or the general decoder (i.e., $P(y_j=w|x,y_{1:j-1})$).
The reason is the same for \texttt{number} in the housing domain and also \texttt{type} in blocks.
The most interesting part is ,\texttt{sydney} where there is a typo in the source sequence in terms of tokenization. 
The caching functions provided the original version of this word in the logical form side.

\section{Examples of caching functions}
Generally the caching function should work as good as the local copying version if we define robust reset functions.
Table \ref{tab:f_history} shows a number of caching functions with different non-linearity ($\sigma$, tanh, or nothing) or different reset gates.
Different functions hold different characteristics and sometimes work better than the proposed function in the paper.
It's worth noting that $f_4,f_5$ provide minimum number of parameters to the model while keeping the token-level performance.
GEOQUERY-S is a variation of the GEOQUERY dataset that we replaced logic forms like $_loc$ with their corresponding natural language forms like $location$; 
we are not able to run these questions in the database though.
\begin{table}[hp]
\centering
\caption{Different models for the caching function.}
\scalebox{0.75}{
\begin{tabular}{ll} \hline
1&$f_1(s_j,c_j) = s_j^TW^{a}W^{(h)}$ \\ \hline
2&$f_2(s_j,c_j,\sigma) = \sigma(s_j^TW^{a}W^{(h)})$ \\ \hline
3&$f_3(s_j,c_j,\sigma) = \sigma(s_j^TW^{a}W^{(h)}+c_jW^{(h)})$ \\ \hline
4&$f_4(s_j,c_j,\sigma) = \sigma(s_j^TU_{z_t})$ \\ \hline
5&$f_5(s_j,c_j,\tanh) = \tanh(s_j^TU_{z_t})$ \\ \hline
6&\specialcell{$f_6(s_j,c_j,\sigma) = z_t \odot \sigma(s_j^TW^{a}W^{(h)}) +{(1-z_t)}\odot$
$ \sigma(c_jW^{(h)})$} \\ \hline
\end{tabular}
}
\label{tab:f_history}
\end{table}

\begin{table}[h]
\small
\caption{Accuracy of different caching functions.}
\scalebox{0.99}{
\begin{tabularx}{\linewidth}{lllll}
\cline{1-5}
    \multirow{2}{*}{ID}&\multicolumn{2}{c}{GEOQUERY}  & \multicolumn{2}{c}{GEOQUERY-S}  \\ \cline{2-5} 
     &SEQ & TOK & SEQ & TOK  \\ \cline{1-5}
     attn&0.771&0.8826 &0.580& 0.8648\\ \cline{1-5}
     $f_1(s_j,c_j)$& \textbf{0.775} & \textbf{0.901} & 0.700 & 0.886
     \\ \cline{1-5}
     $f_2(s_j,c_j,\sigma)$&0.746 & 0.897& 0.711 & 0.88
   \\ \cline{1-5}
     $f_3(s_j,c_j,\sigma)$&0.742&0.893& 0.714& 0.883
 \\ \cline{1-5}
     $f_4(s_j,c_j,\sigma)$& 0.746&0.897 
     &\textbf{0.732} & 0.880\\ \cline{1-5}
     $f_5(s_j,c_j,\tanh)$&0.746 &0.897&0.721& \textbf{0.888}\\ \cline{1-5}
     $f_6(s_j,c_j,\sigma)$& 0.746& 0.8968&\textbf{0.732}&0.883\\ \cline{1-5}
\cline{1-5}
\end{tabularx}
}
\label{tab:res_table:seq2logic}
\end{table}

\end{document}